\documentclass{ecai} 

\usepackage{latexsym}
\usepackage{amssymb}
\usepackage{amsmath}
\usepackage{amsthm}
\usepackage{booktabs}
\usepackage{enumitem}
\usepackage{graphicx}
\usepackage{color}
\usepackage{multirow}

\usepackage{booktabs}

\usepackage{multicol}
\usepackage{makecell}
\usepackage[caption=false]{subfig}

\newcommand{\distas}[1]{\mathbin{\overset{#1}{\kern\z@\sim}}}%

\newcommand{\BibTeX}{B\kern-.05em{\sc i\kern-.025em b}\kern-.08em\TeX}
\definecolor{MyDarkGreen}{rgb}{0.02,0.6,0.02}

\begin{document}


\begin{frontmatter}


\paperid{209} 


\title{SUBLLM: A Novel Efficient Architecture with Token Sequence Subsampling for LLM}




\author[A]{\fnms{Quandong}~\snm{Wang}\thanks{Corresponding Author. Email: wangquandong@xiaomi.com.}\footnote{Equal contribution.}}
\author[B]{\fnms{Yuxuan}~\snm{Yuan}\footnotemark\footnote{Work was done when interning at Xiaomi AI Lab.}}
\author[A]{\fnms{Xiaoyu}~\snm{Yang}}
\author[C]{\fnms{Ruike}~\snm{Zhang}\footnotemark}
\author[A]{\fnms{Kang}~\snm{Zhao}}
\author[A]{\fnms{Wei}~\snm{Liu}}
\author[A]{\fnms{Jian}~\snm{Luan}}
\author[A]{\fnms{Daniel}~\snm{Povey}}
\author[A]{\fnms{Bin}~\snm{Wang}}

\address[A]{Xiaomi AI Lab, China}
\address[B]{Department of Artificial Intelligence, School of Informatics, Xiamen University, China}
\address[C]{Institute of Automation, Chinese Academy of Sciences}


\begin{abstract}
While Large Language Models (LLMs) have achieved remarkable success in various fields, the efficiency of training and inference remains a major challenge. 
To address this issue, we propose SUBLLM, short for Subsampling-Upsampling-Bypass Large Language Model, an innovative architecture that extends the core decoder-only framework by incorporating subsampling, upsampling, and bypass modules. The subsampling modules are responsible for shortening the sequence, while the upsampling modules restore the sequence length, and the bypass modules enhance convergence.
In comparison to LLaMA, the proposed SUBLLM exhibits significant enhancements in both training and inference speeds as well as memory usage, while maintaining competitive few-shot performance.
During training, SUBLLM increases speeds by 26\% and cuts memory by 10GB per GPU. In inference, it boosts speeds by up to 37\% and reduces memory by 1GB per GPU. The training and inference speeds can be enhanced by 34\% and 52\% respectively when the context window is expanded to 8192. Our code is available at https://github.com/XiaoMi/subllm.

\end{abstract}
\end{frontmatter}


\section{Introduction}
Recently in NLP field, the emergence of large language models (LLMs) marks a pivotal advancement in how machines understand and generate human language \cite{brown2020language, JMLR:v21:20-074, touvron2023llama}. Pretrained with huge amounts of parameters on extensive data, LLMs gain extraordinary capabilities across a series of downstream tasks.

Though exhibiting remarkable potential in handling complex tasks, LLMs encounter challenges during training and inference. Firstly, the training process is extremely time-consuming, necessitating the processing of vast amounts of data. Secondly, they need a large amount of GPU memory and computational resources. These factors pose a challenge to their widespread deployment \cite{tay2022efficient, zhu2023survey}.

To address these issues, several approaches have been proposed to accelerate inference and reduce computational costs. Techniques such as distillation \cite{hinton2015distilling, magister2023teaching}, quantization \cite{frantar2022gptq, xiao2023smoothquant}, and pruning \cite{frantar2023sparsegpt, shen-etal-2024-pruning}are employed, as well as decoding optimization \cite{leviathan2023fast, spector2023accelerating} and conditional computation \cite{bapna2020controlling, NEURIPS2023_19d7204a}. 
Additionally, many efforts are dedicated to training acceleration which often leads to inference acceleration too. Some focus on reducing text redundancy \cite{ainslie2023colt5, dai2020funnel, he2023fourier, raposo2024mixture} while others tackle the quadratic computational complexity of the Transformer's self-attention mechanism by improving the attention mechanism \cite{wang2020linformer, zhai2021attention} and proposing new architectures \cite{botev2024recurrentgemma,gu2023mamba, ma2024megalodon,peng2023rwkv}.

Drawing from pertinent research \cite{ainslie2023colt5}, natural language tokens in a sequence vary in importance.
Selectively identifying and removing less significant tokens can significantly reduce computational demands.
Moreover, this targeted approach to prioritizing key information has the potential to enhance training stability, accelerate convergence, and improve overall modeling performance \cite{raposo2024mixture}.

In this paper, we propose a novel and efficient architecture \textbf{S}ubsampling-\textbf{U}psampling-\textbf{B}ypass \textbf{L}arge \textbf{L}anguage \textbf{M}odel (SUBLLM), inheriting the structure of decoder-only LLM, which dynamically allocates computational resources for tokens according to their importance. SUBLLM integrates subsampling and upsampling modules symmetrically between the Transformer blocks, reducing the computational cost while preserving the input sequence's semantics. Specifically, in the subsampling module, a scoring layer calculates each token's importance as the criterion for token subsampling. Meanwhile, a balancer is adopted to adjust the distribution of the token-level score during training. Subsequently, the upsampling module recovers the subsampled sequences to their prior lengths for token prediction in language modeling. Moreover, to improve training stability and accelerate convergence speed, SUBLLM integrates a bypass module that performs a weighted sum of the upsampled token sequence and the original one. The experimental results compared with LLaMA \cite{touvron2023llama} demonstrate the effectiveness of our proposed SUBLLM on model efficiency as well as performance maintenance. The main contributions of this work are summarized as follows: 
\begin{itemize}
\item We propose a novel architecture, SUBLLM, which incorporates subsampling, upsampling, and a bypass module. This design dynamically allocates resources to important tokens, reducing computational costs associated with token redundancy and accelerating model convergence through the bypass connection.

\item We propose a novel approach to token sequence subsampling that effectively measures token importance scores and controls the distribution of score values as expected, thereby achieving the desired subsampling retention ratio during inference.

\item Experimental results demonstrate that SUBLLM achieves 26\% and 37\% speed-up on training and inference respectively compared to the LLaMA model, with a significant reduction of memory cost, while maintaining the performance.
\end{itemize}


\section{Related Work}
\subsection{Training Acceleration}
To reduce the computational cost in LLM training, a lot of work has been carried out from the perspective of reducing redundancy in text. Funnel-Transformer \cite{dai2020funnel} uses strided mean pooling to gradually compress the sequence of hidden states in self-attention, and Fourier-Transformer \cite{he2023fourier} progressively subsamples hidden states with the Fast Fourier Transform operator. 
Same to these methods, our proposed SUBLLM subsamples the tokens to a shortened sequence.
Unlike these methods that reconstruct a sequence by repeating the reduced sequence and adding it back to the original for the final result, SUBLLM's upsampling module takes a different approach. It interpolates between the original and subsampled sequences using token scores as weights, offering a more refined handling of sequence information.

Some other work leverages conditional computation to dynamically expend resources when needed. CoLT5 \cite{ainslie2023colt5} uses conditional routing to decide whether a given token passes through a light branch or a heavy branch in feedforward and attention layers, so as to devote more resources to important tokens. Further, MoD \cite{raposo2024mixture} utilizes a static compute budget, using a per-block router to select tokens for computations, and optimizing FLOP usage by choosing between self-attention and MLP blocks or a residual connection. Our method can also be seen as conditional computation, which dynamically allocates the computational resources to tokens tailored to their importance.  

Another type of work focuses on solving the inefficiency problem caused by the attention mechanism when transformers process sequences. 
Some works focus on improving the attention mechanism to increase the training efficiency \cite{wang2020linformer, zhai2021attention}. 
More recent efforts have introduced novel model architectures to overcome this limitation. 
RWKV \cite{peng2023rwkv} addresses limitations in Transformers by replacing quadratic QK attention with a linear scalar formulation.
RecurrentGemma \cite{botev2024recurrentgemma} combines linear recurrences with local attention to achieve high performance on language tasks with reduced memory usage and faster inference on long sequences. 
Mamba \cite{gu2023mamba} introduces selective state space models, allowing the model to filter out irrelevant information based on the input, enhancing content-based reasoning. 
MEGALODON \cite{ma2024megalodon} introduces an enhanced MEGA architecture with several novel technical components designed to improve capability, efficiency, and scalability. While MEGALODON accelerates training for long sequence inputs with a 32K window length, its training speed at a 4K window length is slower than that of LLaMA. 
In contrast, our proposed SUBLLM builds upon the LLaMA model structure by incorporating subsampling modules to reduce sequence length. As a result, SUBLLM surpasses LLaMA in training efficiency, starting from a window length of 2K.

\subsection{Inference Acceleration}
The parameters that define LLMs are both vast and complex, leading to an ever-increasing demand for computational power and memory capacity. To address these challenges, prior research has primarily concentrated on developing more lightweight models derived from their heavier pre-trained counterparts. 
Key techniques employed in this endeavor include knowledge distillation \cite{hinton2015distilling,magister2023teaching}, quantization \cite{frantar2022gptq, xiao2023smoothquant} and pruning \cite{frantar2023sparsegpt, shen-etal-2024-pruning}. 
However, these methods usually seek the balance between effect and inference acceleration at the cost of sacrificing model performance. 
While the novel model structure SUBLLM we proposed can not only accelerate the inference process but also maintain competitive performance. 

Beyond the sheer scale of LLMs, a significant challenge impacting inference speed is the autoregressive decoding process. The language model decodes text sequentially, requiring K serial iterations to generate K tokens. 
This step-by-step processing not only delays the response time but also turns into a major bottleneck because of the limitations of memory bandwidth. Numerous efforts have been made to optimize the decoding process. 
For instance, speculative decoding \cite{leviathan2023fast,miao2023specinfer,spector2023accelerating} samples from multiple tokens generate from efficient approximation model concurrently as speculative prefixes for the large model. 
LLMA \cite{yang2023inference} accelerates the decoding process by identifying and utilizing overlapping text spans between the LLM output and the reference document.
Medusa \cite{cai2024medusa} expands the model's predictive capabilities through additional heads and a specific tree-based attention mechanism.
MInference \cite{jiang2024minference} accelerates the pre-filling stage by leveraging dynamic sparse attention with spatial aggregation patterns.
YOCO \cite{sun2024yoco} optimizes inference efficiency by reusing global KV caches through cross-attention.
Reducing the KV cache is also an effective method for accelerating inference \cite{liu2023scissorhands,nawrot2024dmc,zhang2023ho}.

In this work, our proposed new architecture SUBLLM is not mutually exclusive with the inference acceleration method above. On the contrary, SUBLLM can also leverage the previously mentioned strategies to expedite the inference process and reduce memory cost. 


\section{SUBLLM}

\begin{figure*}[htbp]
	\centering
        \includegraphics[width=16.5cm]{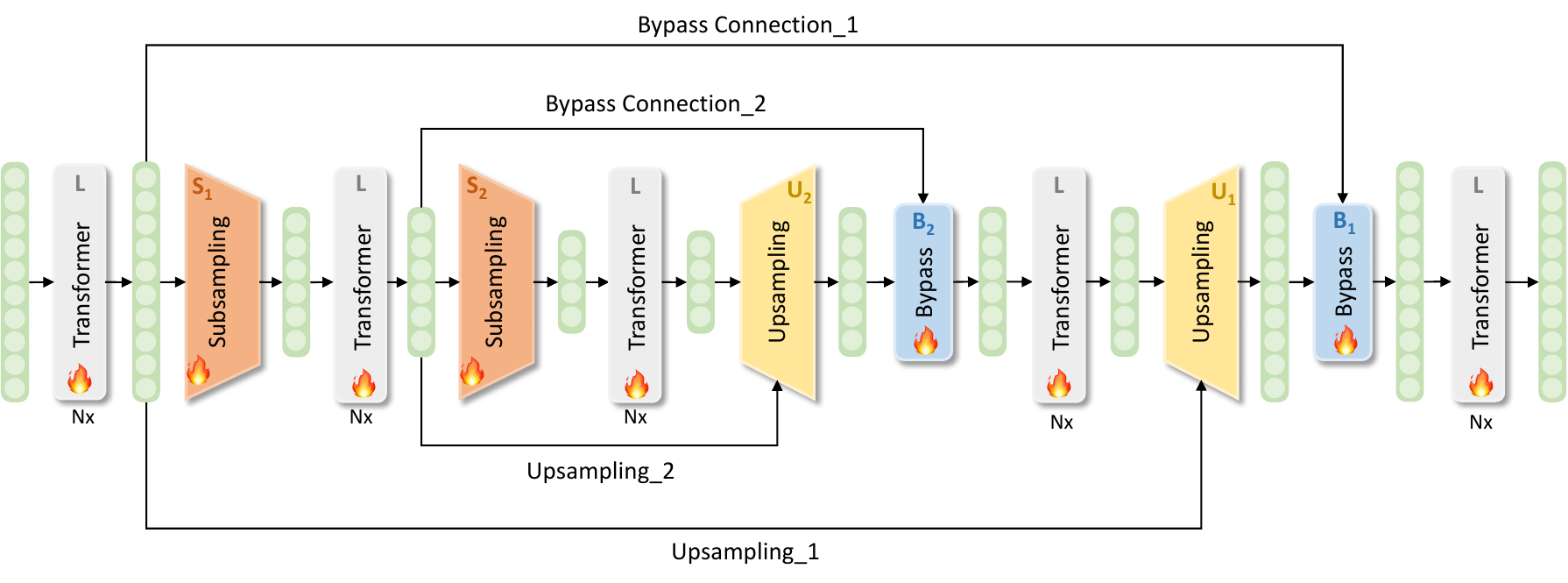}
	\caption{The overall architecture of SUBLLM.}
        \vspace{5mm}
	\label{fig:main}
\end{figure*}

As illustrated in Figure \ref{fig:main}, the proposed SUBLLM model is based on the decoder-only LLM architecture. To manage the number of tokens processed, subsampling and upsampling modules are integrated into the Transformer blocks. The operation proceeds as follows: Initially, the model uses several Transformer blocks to process the full sequence, capturing a comprehensive token sequence representation. Subsampling modules are then introduced, which sequentially eliminate the less critical tokens, thereby reducing the sequence length required for processing. The highest level of sequence compression occurs in the network's middle blocks. Subsequent to this, upsampling modules are employed to reinstate the sequence length. These modules merge the shorter, processed sequences with the original sequences before subsampling, restoring them to their full lengths.
This mechanism allows the decoder-only model to operate as a language model, generating tokens sequentially, which is characteristic of language models where the input and output sequence lengths are identical.
Additionally, we have incorporated bypass connection modules after the upsampling process to utilize each pre-subsampling embedding, assisting to improve the learning process from subsampling to upsampling. Our subsequent experiments confirm that this approach significantly improves convergence efficiency.

For better understanding, we here explain the detailed configurations of SUBLLM. As shown in Table~\ref{tab:SUBLLM_structure}, let $L$ be the Transformer block, $S_i$, $U_i$ and $B_i$ are the corresponding subsampling, upsampling module, and bypass module resepectively. The number before $L$ indicates the number of consecutive Transformer blocks.
Based on the times of subsampling, we evenly divide the number of blocks in the model following the principles of symmetry and minimizing variance between different parts. 
Taking a model with 24 blocks as an example, we strategically place subsampling modules after the outputs of the 5th and 10th blocks, and subsequently put upsampling modules and bypass modules after the outputs of the 14th and 19th blocks. 

\begin{table}[]
    \caption{The structure of the SUBLLM model is represented with a string. The subsample, upsample and bypass module with the same indexes are paired.}
    \centering
    \vspace{3mm}
    \begin{tabular}{ccc}
    \toprule[1pt]
    Blocks & S/U Num & Model representation \\
    \midrule
    15 & 1  & $5L\_S_1\_5L\_U_1\_B_1\_5L$ \\
    15 & 2  & $3L\_S_1\_3L\_S_2\_3L\_U_2\_B_2\_3L\_U_1\_B_1\_3L$ \\
    24 & 2  & $5L\_S_1\_5L\_S_2\_4L\_U_2\_B_2\_5L\_U_1\_B_1\_5L$ \\
    \bottomrule[1pt]
    \end{tabular}
    \label{tab:SUBLLM_structure}
\end{table}

\subsection{Learnable Subsampling Module}
The subsampling module consists of a scoring layer and an activation balancer.
Given an input token sequence $\textbf{x}=\textbf{x}_1, \dots, \textbf{x}_N$ of length $N$, the subsampling module reduces the sequence length by discarding redundant tokens. 
Let the subsampling retention ratio be $d$ ($d$ smaller than 1), the subsampled sequence $\textbf{x}'$ is 
\begin{align}
    \textbf{x}' &=\textsc{IndexSelect}(\textbf{x}, \mathcal{I}) \\
     &= x_{\mathcal{I}_1},\dots, x_{\mathcal{I}_{N'}}
\end{align}

where $\mathcal{I}$ is the set of indexes of the kept tokens after subsampling and $N'=\lceil N * d \rceil$ is the length of the subsampled sequence. \\

\noindent \textbf{Token Selection} The indexes of the kept tokens are determined by their importance values. To evaluate the importance of each token, a scoring layer predicts the token-level scalar importance value $\textbf{w}=w_1, \dots, w_N$, also known as weight:
\begin{align}
    \textbf{s} &= s_1, \dots, s_N \\
     & =  \mathcal{S}(\textbf{x}_1), \dots, \mathcal{S}(\textbf{x}_N)\\
    w_n &= \textsc{Clamp}(\textsc{Balancer}(s_n, [0,1]))
    \label{eqn:weight}
\end{align}

where $\mathcal{S}$ is the scoring layer, $\textbf{s}$ is the token-level score, $\textsc{Clamp}$ is the operation clamping the $n$-th token's score value $s_n$ to $[0,1]$ and $\textsc{Balancer}$ is a balancer module controlling the distribution of the $\textbf{s}$. As the same to the decoder-only language model, the scoring layer should not rely on the future tokens. In this work, $\mathcal{S}$ adopts the simplest structure: a single linear layer mapping the dimension of Transformer embedding to a scalar value. During training, the kept indexes $\mathcal{I}$ and discarded indexes $\hat{\mathcal{I}}$ can be formulated as:
\begin{align}
    \mathcal{I} &= \{i | w_{i} \in \textsc{TopK}(\textbf{w}, N')\} \\
    \hat{\mathcal{I}} &= \{i | w_{i} \not \in \textsc{TopK}(\textbf{w}, N')\}
    \label{eqn:topK}
\end{align}
where $\textsc{TopK}(\textbf{w}, N')$ stands for the top $N'$ values among $\textbf{w}$. We keep the original sequential order after subsampling, i.e. the elements in $\mathcal{I}$ are sorted. 
Note that the $\textsc{TopK}$ operation is cumbersome to implement during inference as previously discarded tokens can be among the top-K tokens after the language model has emitted a few less important tokens. A different token selection strategy is employed in the inference mode to circumvent this problem, which will be discussed in the later section.

\noindent\textbf{Positional Encoding Subsampling}
We use RoPE \cite{su2024roformer} for relative positional encoding. After subsampling, tokens that were originally distant might become adjacent, distorting the positional encoding if the subsampled sequence is treated as a new sequence. To address this, we store the indexes of the retained tokens, $\mathcal{I}$, and use them to subsample the sine and cosine matrices used in the RoPE module along the sequence dimension. This ensures that the relative positional information in the subsampled sequence remains consistent with the original sequence.
The relative positional encoding of a token pair $(i,j)$ in the subsampled sequence $\textbf{x}'$ is formulated as follows:
\begin{align}
    RelPos(\textbf{x}'_i, \textbf{x}'_j) = \textsc{RoPE}(\textbf{x}'_i, \textbf{x}'_j, (\mathcal{I}_i - \mathcal{I}_j))
\end{align}
where $\textsc{RoPE}$ stands for the RoPE module in LLaMA family models.

\noindent \textbf{Inference Mode} As mentioned earlier, it is impossible to apply Equation~\ref{eqn:topK} to select the important tokens as the entire token sequence is not available due to the auto-regressive nature of the language model. 
To tackle this issue, an approximation can be applied to obtain the indexes of the kept tokens:
\begin{align}
    \mathcal{I}_{infer} &= \{i | w_{i}  \geq v \} 
    \label{eqn:inference}
\end{align}
where $v$ is a hyper-parameter between 0 and 1. Tuning $v$ is equivalent to adjusting the balance between the actual inference speed and the model accuracy. The larger $v$ is, the less tokens are kept after subsampling, leading to a faster inference speed. An ideal value of $v$ is supposed to make the actual subsampling retention ratio during inference as close as possible to $d$.

\noindent\textbf{Balancer}
Due to the approximation applied in Equation~\ref{eqn:inference}, the kept proportion of tokens is changing dynamically and can differ from $d$. To reduce the gap between training and inference, it is important to keep the proportion of $\textbf{w} \geq v$ close to $d$.
To encourage this behavior, a balancer \cite{yao2023zipformer} module is added before the clamping operation to control the maximum/minimum proportion of positive values of $\textbf{s}$, denoted as $p_{max}$ and $p_{min}$, and the upper/lower bound of the mean absolute weight values of $\textbf{s}$, denoted as $a_{max}$ and $a_{min}$. 
This is achieved by adding an extra gradient term enforcing the distribution of $\textbf{s}$ to the gradient back-propagated to $\textbf{s}$. 
Suppose the subsampling retention ratio is $d$, the setting of the balancer module is as follows:
\begin{align}
     p_{max} &=d + 0.05, p_{min} =d - 0.05 \\
     a_{max} &=4.0, a_{min}=1.0
\end{align}
Intuitively, the balancer limits the proportion of positive score values to the range $d \pm 0.05$ during training. In this case, using $v=0$ as threshold in Equation~\ref{eqn:inference} makes the inference behavior close to the training setting, which is also adopted in out final implementation. Limiting the upper bound of the score value ($a_{max}=4$) before clamping prevents a too-large and too-sparse gradient during back-propagation. Note that the balancer does not have learnable parameters and is only activated during training, so $\textbf{s}$ is untouched during inference. 

\subsection{Upsampling Module}
The upsampling module reconstructs a subsampled token sequence to its original length prior to subsampling.
Let $S_n$ be a subsampler subsampling $\textbf{x}$ of length $N$ to $\textbf{x}'$ of length $N'$, its paired upsampler $U_{n}$ utilizes the token indexes $\mathcal{I}$ of $\textbf{x}'$, the token-level weights $\textbf{w}$ of $\textbf{x}$ and token sequence $\textbf{x}$ to produce a new token sequence $\textbf{x}_{new}$ of length N following the procedures below. First, a token-level scaling factor $\textbf{w}_{scaling}$ of length $N'$ is computed:
\begin{align}
    \textbf{w}_{kept} &= \textsc{IndexSelect}(\textbf{w}, \mathcal{I}) \label{eq:index_select}\\
    \textbf{w}_{discarded} &= \textsc{IndexSelect}(\textbf{w}, \hat{\mathcal{I}}) \\
    \textbf{w}_{sample_i} &\sim \textsc{Uniform}(\mathcal{E}), \; \text{for} \; i=1,2,...,N'\\
    \textbf{w}_{scaling} &= \textbf{w}_{kept} - \textbf{w}_{sample} \label{eq:weight_scaling}
\end{align}
where $\mathcal{E}$ is the set of the elements in $\textbf{w}_{discarded}$. This enables arbitrary subsampling rate as the length of $\textbf{w}_{kept}$ and $\textbf{w}_{discarded}$ are unnecessarily the same. $\textbf{w}_{scaling}$ serves as a scaling factor when constructing the upsampled sequence $\textbf{x}_{new}$:
\begin{align}
    \textbf{x}_{new, i} = 
    \begin{cases}
    \textbf{w}_{scaling, i} * \mathcal{G}(\textbf{x}_i) + (1- \textbf{w}_{scaling, i}) * \textbf{x}_i,& \text{if } i \in \mathcal{I}\\
     \textbf{x}_i,              & \text{otherwise}
\end{cases}
\end{align}
where $\mathcal{G}$ represents the intermediate transformations that $\textbf{x}_i$ goes through after the subsampler $S_n$ and before the upsampler $U_n$ (e.g. several Transformer blocks or other nested down/upsamplers). 
Note that instead of directly using $\textbf{w}_{kept}$, we employed $\textbf{w}_{scaling}$ to scale $\mathcal{G}(\textbf{x}_{i})$, which is obtained by subtracting the weights of the discarded tokens from the kept tokens. The motivation for this strategy is twofold. First, the subtraction makes the selection of the tokens fully differentiable as the gradient associated with the discarded tokens can be back-propagated. Second, the scoring layer in the subsampler could learn to emit a large score (i.e. 1 after clamping) for all tokens without a penalizing measure. Using $\textbf{w}_{scaling}$ discourages the model from this behaviour by penalizing the weight values of the discarded tokens, promoting the model to discriminate between more important and less important tokens. 
The reconstructed sequence $\textbf{x}_{new}$ can be interpreted an interpolation between the subsampled sequence and the original token sequence before subsampling. As a result, the discarded tokens go through fewer Transformer blocks than the kept tokens. In extreme cases where $\textbf{w}_{i}=1$ for all $i\in\mathcal{I}$, and $\textbf{w}_{i}=0$ for all $i\in\hat{\mathcal{I}}$, the upsampled token sequence is just a index-level re-ordering of $\mathcal{G}(\textbf{x})$ and $\textbf{x}$ as the original token order. 

\subsection{Bypass Module}

A bypass module is added to combine the output $\textbf{y}$ of a group of modules with its input $\textbf{x}$. It learns a channel-wise weight $\textbf{c} \in \mathbb{R}^{C}$ between $[0,1]$ to control the throughput of each Transformer block:
\begin{align}
    \textbf{y} =(1-\textbf{c}) \odot \textbf{x} + \textbf{c} \odot \textbf{y}
\end{align}
where $C$ is the feature dimension of $\textbf{y}$ and $\odot$ represents the channel-wise multiplication. A larger $\textbf{c}$ makes the model ``straight-through'' by increasing the contribution of $\textbf{y}$. In SUBLLM, one bypass module is added to each paired subsample/upsample modules, i.e. combining the input of $S_i$ with the output of $U_i$. Bypass module accelerates the convergence of SUBLLM by enforcing all layers to learn high-quality representations, especially at the beginning of the training. 
A value range can be applied to limit all entries $\textbf{c}_j$ of $\textbf{c}$ to the range $[c_{min}, c_{max}]$. This is achieved by negating the positive gradient w.r.t to $\textbf{c}_j$ if $\textbf{c}_j$ is smaller than $c_{min}$, or negating the negative gradient w.r.t to $\textbf{c}_j$ if $\textbf{c}_j$ is bigger than $c_{max}$.



\section{Experiments}

\begin{table*}[htbp]
\caption{Experimental results of LLaMA baseline and our proposed SUBLLM on computational efficiency and performances.  For the evaluation of computational efficiency, speed-up and memory reduction during both pre-training and inference serve as metrics. TGS represents the number of processing tokens per GPU per second. Inference speed is the number of processed tokens per second. For the evaluation of model performance, we consider valid loss during pre-training and few-shot learning in downstream tasks.}
\vspace{4mm}
\centering
\begin{tabular}{lcccccccc}
    \toprule[1pt]
    
    \multirow{3}*{\makecell[l]{\textbf{Efficiency}}} & \multicolumn{5}{c}{\textbf{Pre-Training}} & \multicolumn{3}{c}{\multirow{2}*{\textbf{Inference}}} \\
    & \multicolumn{3}{c}{Speed-Up} & \multicolumn{2}{c}{Max Speed-Up} &  \\	
    \cmidrule(rl){2-4}  \cmidrule(rl){5-6} \cmidrule(rl){7-9} 
    & TGS$\uparrow$ & Ratio$\uparrow$ &  Mem (GB)$\downarrow$ & TGS$\uparrow$ & Ratio$\uparrow$  & \makecell[c]{Speed$\uparrow$} & \makecell[c]{Ratio$\uparrow$} & \makecell[c]{Mem (GB)$\downarrow$} \\
    \hline
    \textbf{LLaMA}    & 16,976 & $\times$1.00 & 65.99 & 18,856 & $\times$1.00 & 17.83 & $\times$1.00 & 18.49   \\
    \textbf{SUBLLM} & \textbf{21,341} & \textbf{$\times$1.26} & \textbf{55.81} & \textbf{24,773} & \textbf{$\times$1.31} & \textbf{24.43} & \textbf{$\times$1.37} & \textbf{17.29} \\

    \toprule[1pt]
    \multirow{2}*{\makecell[l]{\textbf{Performance}}} & \textbf{Pre-Training} & \multicolumn{7}{c}{\textbf{Few-Shot Learning}} \\
    \cmidrule(rl){2-2}  \cmidrule(rl){3-9} 
    & Valid Loss$\downarrow$ & SST2$\uparrow$ & Amazon$\uparrow$ & DBpedia$\uparrow$ & AGNews$\uparrow$ & Yelp$\uparrow$ & Hate$\uparrow$ & Avg.$\uparrow$ \\
    \hline 
    \textbf{LLaMA}    & 3.11 & 81.01 & 86.54 & \textbf{45.70}	& 64.77 & 87.59 & \textbf{45.18} & 68.47 \\
    \textbf{SUBLLM} & \textbf{3.10} & \textbf{91.95} & \textbf{94.57} & 42.97 & \textbf{66.05} & \textbf{94.24} & 32.23 & \textbf{70.34} \\
    \bottomrule[1pt]
\end{tabular}
\label{tab:main}
\end{table*}

\subsection{Settings}

\paragraph{Pre-Training Corpora} 
We use SlimPajama \cite{cerebras2023slimpajama} as the pre-training corpus, which includes CommonCrawl, C4, Wikipedia, GitHub, StackExchange, ArXiv, and Book datasets, sampled according to SlimPajama's original proportions. 

\paragraph{Pre-Training Details} 
We adopt LLaMA2 \cite{touvron2023llama2} as the baseline, training 1.3B \cite{xia2024shearedllama} and 0.25B parameter versions. 
SUBLLM shares the same training configuration but introduces only 8,192 additional parameters for the 1.3B model and 4,096 for the 0.25B model.
Each model is trained with 100 times the number of its parameters in tokens and has versions with context window sizes of 2K, 4K, and 8K. 
Eden \cite{yao2023zipformer} is used for the learning rate schedule, ScaledAdam \cite{yao2023zipformer} as the optimizer, and ZeRO \cite{rajbhandari2020zero} to enhance training efficiency and optimize resource utilization.
Training is conducted with bf16 precision using Fairseq \cite{ott2019fairseq}, with Flash Attention \cite{dao2023flashattention} to accelerate the process.
More details are in the supplementary material \cite{wang2024subllm}.

\subsection{Main Results}
Table \ref{tab:main} provides the experimental results of LLaMA and the proposed SUBLLM on the computational efficiency during the pre-training and inference phases, as well as model performance. Both models are with the same configuration of 1.3B model size and 4K context window. The minimal retention ratio of the input tokens in SUBLLM subsampling is 40\%, which will be discussed in detail in the following section.

\paragraph{Computational Efficiency}

We explore the computational resource savings of our model, specifically focusing on training and inference acceleration as well as GPU memory reduction. Pre-training speed-up, which is evaluated in the number of tokens that each GPU processes for each second (i.e. TGS), reveals a 26\% increase for SUBLLM compared to LLaMA with the same batch size. Meanwhile, the pre-training of SUBLLM achieves a significant memory reduction of 10GB compared with LLaMA. The improvement in pre-training speed is further enhanced when the memory saved by SUBLLM is reallocated to increase the batch size, boosting the training acceleration from 26\% to 31\%, which we marked as max speed-up.

As for inference acceleration, SUBLLM displays a 37\% increase in speed, higher than the 26\% improvement observed during training. Because in trainig stage, SUBLLM only accelerate the forward and backward process rather other computations like parameters update.
For clarity, the referenced decoding speed specifically pertains to the decoding of non-first tokens on a single GPU. Also, SUBLLM contributes to 1GB memory reduction compared with LLaMA. These results indicate SUBLLM is a valuable architecture for tasks requiring high-level computational efficiency.

\paragraph{Model Performance}

In evaluating the model performance of SUBLLM, we consider both the valid loss during pre-training and its performance on few-shot learning tasks. As shown in Table \ref{tab:main}, SUBLLM's valid loss is on par with that of LLaMA, indicating that its token prediction capabilities are comparable. For few-shot learning, we evaluate SUBLLM on 6 text classification datasets including sentiment classification (SST2 \cite{socher2013recursive}, Amazon and Yelp \cite{zhang2015character}), topic classification (DBpedia \cite{lehmann2015dbpedia}, AGNews \cite{zhang2015character}) and hate speech detection (Hate \cite{barbieri2020tweeteval}). Despite some fluctuations in scores across different datasets, the overall performance of SUBLLM is broadly equivalent to 
LLaMA. Both findings above indicate the validity of the optimized architecture on the model performance in token prediction as well as few-shot in-context learning.
The results of the model on other benchmarks \cite{mmlu2021, suzgun2022bbh, zhong2023agieval} can be found in Table \ref{tab:benchmark_results} in the supplementary materials \cite{wang2024subllm}.

\subsection{Ablation Study}
We perform an ablation study on the 0.25B SUBLLM model to examine the effects of the bypass module on validation loss. The findings summarized in Table \ref{tab:abl} show that SUBLLM with all enhancements achieves the lowest valid loss of 3.66. Changing the bypass module's operation from a weighted sum to a standard residual connection increases the validation loss, which is higher than LLaMA. This results demonstrates the importance of weighted integration. Completely removing the bypass module leads to a further increase in validation loss, which confirms the bypass module's role in maintaining low valid loss by using intermediate token information. Overall, the bypass module significantly enhances learning efficiency.

\begin{table}[t]
\centering
\caption{Ablation results of the proposed bypass module of SUBLLM.}
\vspace{5mm}
\begin{tabular}{lc}
\toprule[1pt]
\textbf{Variant}                                & \textbf{Valid Loss}$\downarrow$  \\ 
\hline
SUBLLM                                         & \textbf{3.66} \\
\enspace - Bypass Module + Residual Connection   & 3.72  \\         
\enspace - Bypass Module                         & 3.73 \\
LLaMA                                            & 3.69 \\
\bottomrule[1pt]
\end{tabular}
\label{tab:abl}
\end{table}

\begin{table*}[htbp]
    \centering
    \caption{Detailed analysis of computational efficiency in the pre-training phase, where the max speed-up reallocates the saved memory in pre-training for a larger batch size to further explore the maximum speed-up boundary.}
    \vspace{4mm}
    \begin{tabular}{cclcccccc}
    \toprule[1pt]
        \multirow{2}*{\makecell[c]{\textbf{Model} \\ \textbf{Size}}} & \multirow{2}*{\makecell[c]{\textbf{Context} \\ \textbf{Window}}} & \multirow{2}*{\textbf{Model}} & \multicolumn{4}{c}{\textbf{Speed-Up}} & \multicolumn{2}{c}{\textbf{Max Speed-Up}}  \\
        \cmidrule(rl){4-7} \cmidrule(rl){8-9} 
        & & & TGS$\uparrow$ & Ratio$\uparrow$ & Mem (GB)$\downarrow$ & $\Delta$ Mem & TGS$\uparrow$ & Ratio$\uparrow$ \\
        \hline 
        \multirow{6}*{0.25B} & \multirow{2}*{2k}  & LLaMA & 85,925 & $\times$1.00 & 60.16 & - & 85,462 & $\times$1.00 \\
                             &                    & SUBLLM & 107,260 & $\times$1.25 & 53.76 &	-6.41  & 107,859 & $\times$1.26 \\
        \cmidrule(rl){2-9}
                             & \multirow{2}*{4k}  & LLaMA & 77,590 & $\times$1.00 & 77.66 &	- & 77,423 & $\times$1.00 \\
                             &                    & SUBLLM & 99,209 & $\times$1.28 & 69.03 &	-8.63 & 100,425 & $\times$1.30 \\
        \cmidrule(rl){2-9}
                             & \multirow{2}*{8k}  & LLaMA & 64,959 & $\times$1.00 & 74.15 & 	- & 64,741 & $\times$1.00 \\
                             &                    & SUBLLM & 86,227 & $\times$1.33 & 66.03 &	-8.11 & 87,261 & $\times$1.35 \\
    \toprule[1pt]
        \multirow{6}*{1.3B}  & \multirow{2}*{2k}  & LLaMA & 18,405 & $\times$1.00 & 72.99 & - & 20,284 & $\times$1.00 \\
                             &                    & SUBLLM & 22,831 & $\times$1.24 & 61.80 & -11.19 & 26,219 & $\times$1.29 \\
        \cmidrule(rl){2-9}
                             & \multirow{2}*{4k}  & LLaMA & 16,976 & $\times$1.00 & 65.99 & 	- & 18,856 & $\times$1.00 \\
                             &                    & SUBLLM & 21,341 & $\times$1.26 & 55.81 &	-10.18 & 24,773 & $\times$1.31 \\
        \cmidrule(rl){2-9}
                             & \multirow{2}*{8k}  & LLaMA & 15,080 & $\times$1.00 & 65.89 & 	- & 16,587 & $\times$1.00  \\
                             &                    & SUBLLM & 19,390 & $\times$1.29 & 56.19 &	-9.70 & 22,264 & $\times$1.34 \\
    \bottomrule[1pt]
    \end{tabular}
\label{tab:pre-training}
\end{table*}

\begin{table*}[htbp]
    \centering
    \caption{Detailed analysis of computational efficiency in the inference phase. Actual retention means the lowest retention rate of the sequence tokens through the depth of the model in the inference phase. }
    \vspace{4mm}
    \begin{tabular}{clcccccccc}
    \toprule[1pt]
         \multirow{2}*{\makecell[c]{\textbf{Context} \\ \textbf{Window}}}& \multirow{2}*{\textbf{Model}} & \multirow{2}*{\makecell[c]{\textbf{Actual} \\ \textbf{Retention}}} & \multicolumn{2}{c}{\textbf{First Token}} & \multicolumn{2}{c}{\textbf{Non-First Tokens}} & \multicolumn{2}{c}{\textbf{Memory}} \\
        \cmidrule(rl){4-5} \cmidrule(rl){6-7} \cmidrule(rl){8-9}
        & & & Latency (ms)$\downarrow$ & Ratio$\uparrow$ & Speed $\uparrow$ & Ratio$\uparrow$ & Mem (GB)$\downarrow$ & $\Delta$ Mem \\
        \hline
        \multirow{2}*{2k} & LLaMA & - & 695.16 & $\times$1.00 & 20.82 & $\times$1.00 & 6.98 & - \\
        & SUBLLM & 43\% & 496.66 & $\times$1.40 & 26.71 & $\times$1.28 & 5.63 & -1.35 \\     
        \hline
        \multirow{2}*{4k} & LLaMA & - & 2,051.59 & $\times$1.00 & 17.83 & $\times$1.00 & 18.49 & -\\
        & SUBLLM & 44\% & 1,410.94 & $\times$1.45 & 24.43 & $\times$1.37 & 17.29 & -1.20  \\
        \hline
        \multirow{2}*{8k} & LLaMA & - & 16,249.11 & $\times$1.00 & 12.38 & $\times$1.00 & 61.05 & - \\
        & SUBLLM & 44\% & 9,758.40 & $\times$1.67 & 18.80 & $\times$1.52 & 58.61 & -2.44  \\
   \bottomrule[1pt]
    \end{tabular}
\label{tab:inference}
\end{table*}

\begin{table}[htbp]
\centering
\caption{The impact of Adam and ScaledAdam optimizers on the model performance and speed-up in pre-training.}
\vspace{5mm}
\begin{tabular}{lcccc}
\toprule[1pt]
& \multicolumn{2}{c}{\textbf{Adam}}  & \multicolumn{2}{c}{\textbf{ScaledAdam}} \\ 
\cmidrule(rl){2-3}  \cmidrule(rl){4-5} 
& Valid Loss$\downarrow$ & Ratio & Valid Loss$\downarrow$ & Ratio \\
\hline 
\textbf{LLaMA}             & 3.725 & - & 3.693 & - \\
\textbf{SUBLLM}          & 3.743 & $\times$1.33 & 3.687 & $\times$1.32 \\
             
\bottomrule[1pt]
\end{tabular}
\label{tab:opt}
\end{table}

\section{Analysis and Discussions}

\subsection{Detailed Analysis of Computational Efficiency}
\subsubsection{Pre-Training}
The experimental results outlined in Table \ref{tab:pre-training} offer a comprehensive comparison of the SUBLLM and LLaMA models, highlighting the improvements in pre-training speed and reductions in memory usage across various configurations. Specifically, the table illustrates the performance metrics for models with sizes of 0.25B and 1.3B. The results for the 0.25B model were obtained using a single node equipped with 8 A100 GPUs. For 1.3B model, the training speed-up were recorded using four nodes, while the max speed-up results were obtained with a single node.

\paragraph{Speed-Up} In the analysis of training speed, the SUBLLM model consistently improves in tokens per GPU per second (TGS) and speed-up ratio as the context window size increases. This shows that SUBLLM is more efficient in larger contexts. Meanwhile, the 1.3B model also shows increased speed-up ratios. Although the increment is slightly less than the 0.25B model, it is likely due to higher communication overhead in multi-node setups. During the training process, the calculation of forward pass and backward propagation is related to the context window. Other calculations including gradient calculation and parameter update have nothing to do with the sequence length. Therefore, as the window length increases, the acceleration effect achieved by subsampling is better, and the acceleration ratio increases.

\paragraph{Max Speed-Up} The right section of Table \ref{tab:pre-training} shows the maximum achievable speed-up, where the reduced memory is allocated for a larger batch size to further accelerate the pre-training process. We can see that as the sequence length increases from 2k to 8k, the speed-up effect compared with LLaMA becomes more significant, and more importantly larger than the regular speed-up ratio in the same batch size.
Noticed that SUBLLM also gains a higher max speed-up ratio as the language model scales up to 1.3B, where the speed-up ratio gap between SUBLLM and LLaMA becomes larger.

\paragraph{Memory} Concerning GPU memory usage, SUBLLM markedly improves upon LLaMA, with the memory savings for the 0.25B model increasing from  6GB in a 2k window to 8GB in an 8k window. This substantial reduction in memory usage with larger window sizes underlines SUBLLM’s enhanced processing efficiency. The 1.3B model mirrors this pattern, confirming the model's improved efficiency in more extensive configurations. Overall, SUBLLM not only boosts training speeds but also significantly reduces the memory footprint, making it especially advantageous in larger configurations. This scalability and efficiency position SUBLLM as an attractive option for environments where optimal performance and effective computational resource management are paramount.

\subsubsection{Inference}

Table~\ref{tab:inference} provides a detailed analysis of the inference acceleration on one A100 GPU and GPU memory savings for the 1.3B SUBLLM model that subsamples sequences twice to retain 40\% of the sequence. The tests covered 1.3B models with 2k, 4k, and 8k window sizes, assessing various performance metrics across corresponding input lengths of 2k, 4k, and 8k. The test samples were taken from the Slimpajama test set, and the inference batch size was set to 8. The metrics evaluated included initial token latency and the acceleration ratio of SUBLLM over LLaMA, non-initial token latency and its acceleration ratio, GPU memory usage during inference, the memory savings achieved by SUBLLM, and the actual token retention ratio during inference.

The results show that the actual token retention ratio during inference slightly exceeds the 40\% set during training, which is expected due to the training balancer allowing for a fluctuation range of $\pm$5\%. As the input sequence length increases, both the initial and subsequent token decoding speeds show greater acceleration ratios. 
This improvement is related to model inference with Pytorch. SUBLLM reduces the computational load and significantly decreases the time spent on computations within the CUDA kernels. However the time required to launch CUDA kernels remains constant across different sequence lengths. With longer sequences, the proportion of time spent within CUDA kernels becomes more significant. Thus, these factors leads to higher speed-up ratios for longer sequences.

Additionally, the observed increase in memory savings correlates with the reduction in the size of key-value cache, which refers to the length of the key-value pairs stored in the cache for the attention mechanism during inference. These findings demonstrate that the SUBLLM model structure yields greater benefits when processing longer texts during inference, highlighting its efficiency and effectiveness in large-scale text handling.

\begin{figure}[t]
    \centering
    \subfloat[Subsampling retention ratio]{\includegraphics[width=0.43\textwidth]{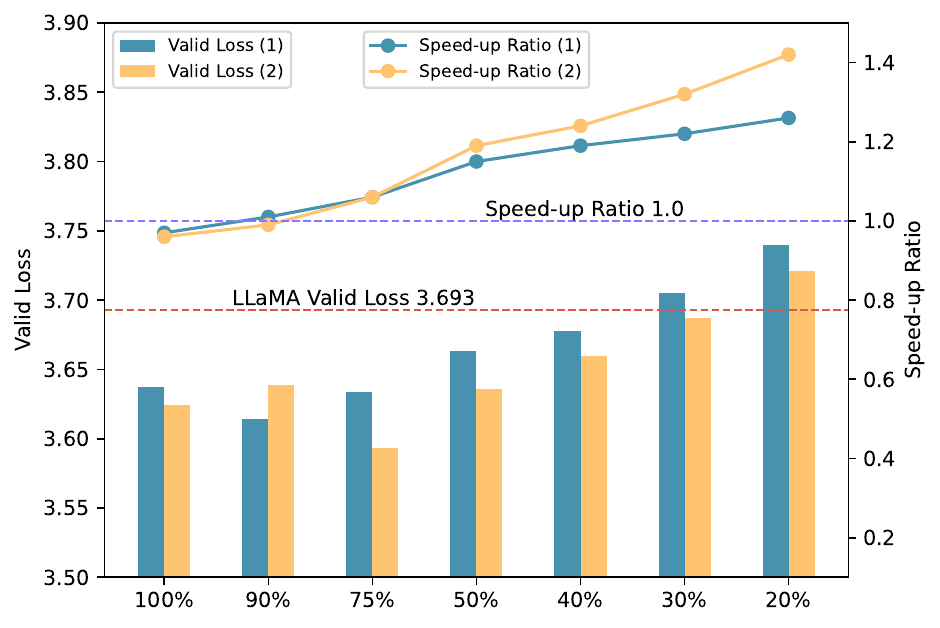}
    \label{fig:sub_r}}
    \hspace{0.5cm}
    \subfloat[The number of subsampling blocks] {\includegraphics[width=0.43\textwidth]{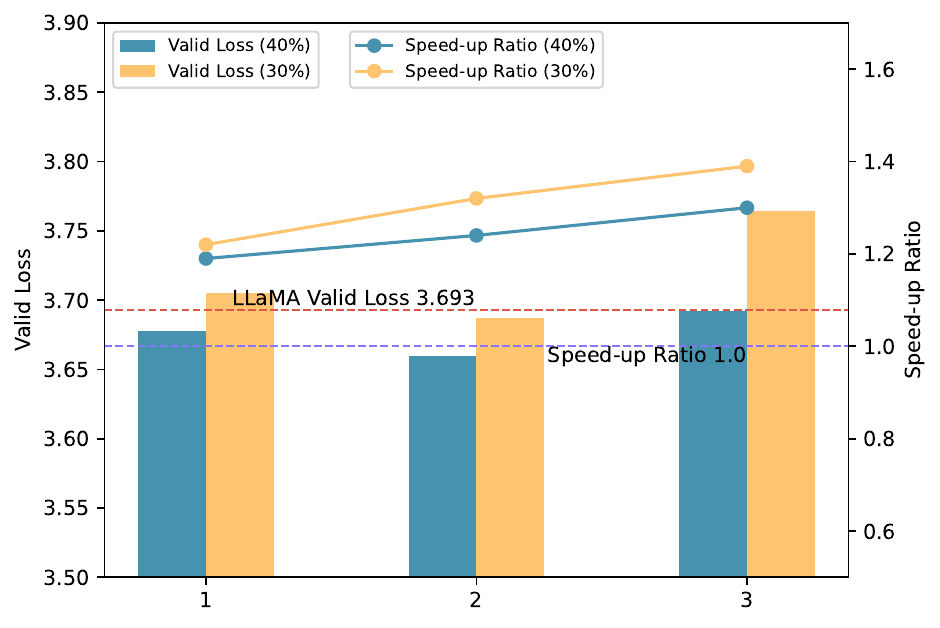}
    \label{fig:sub_t}}
    \caption{The impact of various subsampling setups on model performance and speed-up in pre-training. Figure \ref{fig:sub_r} illustrates the model with one and two subsampling modules, denoted by (1) and (2), respectively.}
    \vspace{5mm}
    \label{fig:subsampling}
\end{figure}

\subsection{Analysis on Subsampling}
We explore the impact of different subsampling setups on the model performance (i.e., valid loss) and training speed-up ratio, including the number of continuous subsampling modules and retention ratio. This retention ratio refers to the lowest retention rate of the original sequence length through the depth of the model. Aiming to search for an optimal configuration with an appropriate speed-up ratio and better performance that can be applied universally, we conduct experiments on the proposed SUBLLM with 0.25B parameters for training efficiency and parameter selection. Note that if the two configurations have close speed-up ratios, we choose the one with better performance as the optimal configuration.

\paragraph{Retention Ratio} From Figure \ref{fig:sub_r} we can see that SUBLLM achieves the lowest valid loss by retaining 90\% of tokens with subsampling once and retaining 75\% of tokens with subsampling twice, yet the speed-up ratio is relatively low around 1.0. When the retention ratio is 40\% and 30\%, the valid loss of SUBLLM is lower than LLaMA and the training speed-up is significant, especially with subsampling twice continuously. In addition, for 100\% retention rate without discarding tokens in pre-training, the valid loss of SUBLLM is still lower than LLaMA, demonstrating the effectiveness of the bypass module of SUBLLM for convergence acceleration and loss reduction.

\paragraph{Subsampling Times}
We further conduct experiments on the variants of subsampling times under the retention ratios of 30\% and 40\%. As shown in Figure \ref{fig:sub_t}, the valid loss of subsampling twice is lower than that of subsampling once. It can also be observed from the figure that subsampling three times leads to the valid loss increase, especially for a 30\% retention ratio. This probably results from that the Transformer blocks are relatively few between paired subsampling modules, which is not sufficient for extracting high-level semantic information for each processed sequence and leads to suboptimal performance. Given the priority of performance optimization, we consider 2 successive subsampling with retaining 40\% tokens as the optimal configuration with a prominent pre-training efficiency.

\subsection{Analysis of Optimizer}

The experimental results presented in Table \ref{tab:opt} analyze the impact of different optimizers on the performance of 0.25B models, specifically focusing on Adam and ScaledAdam. Both LLaMA and SUBLLM are evaluated with valid loss and speed-up ratios during pre-training, where the batch sizes are the same between these two models. Employing ScaledAdam for optimization leads to lower valid losses for both models (especially for SUBLLM), suggesting ScaledAdam could facilitate convergence and improve model performance. SUBLLM achieves a consistent speed-up ratio of 1.33 with Adam and 1.32, indicating that ScaledAdam improves model performance while not hurting computational efficiency. This analysis sets a benchmark for future optimizations and model enhancements.

\subsection{Validity of Subsampling}

\begin{figure}[t]
    \centering
    \subfloat[SUBLLM]{\includegraphics[width=0.45\columnwidth]{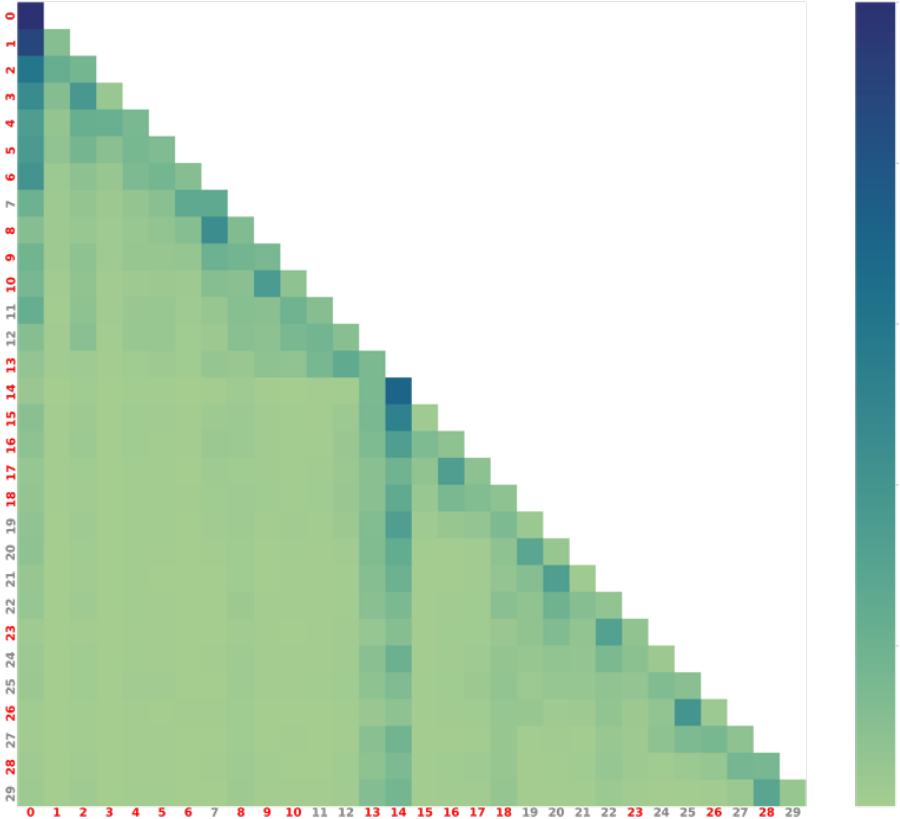}
    \label{fig:att1}}
    \hspace{0.3cm}
    \subfloat[LLaMA]{\includegraphics[width=0.45\columnwidth]{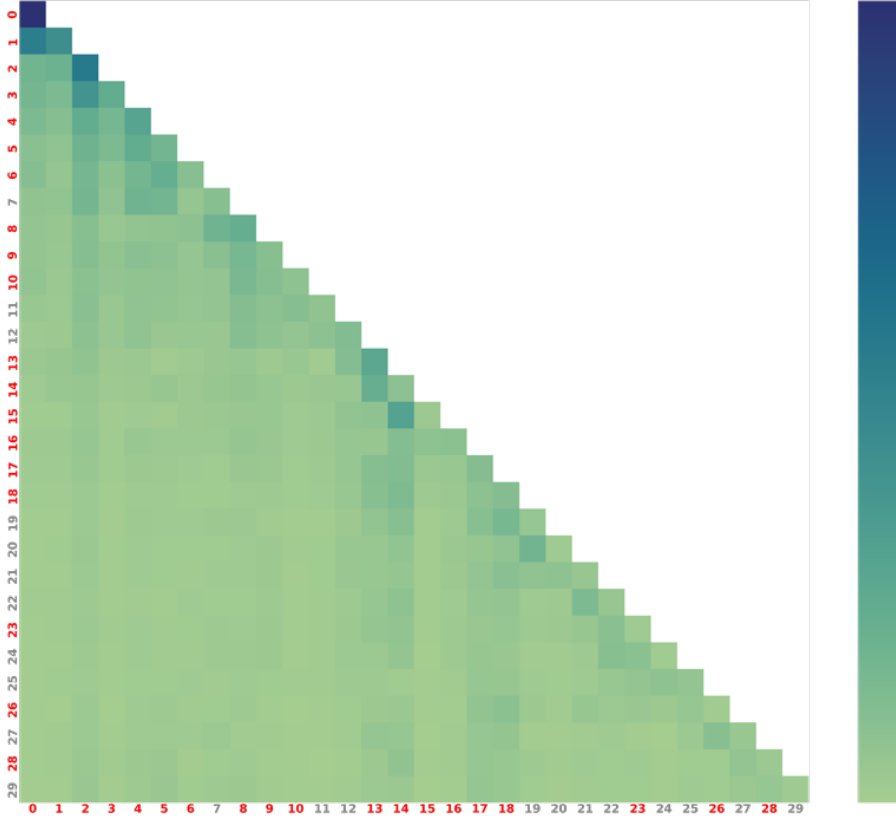}
    \label{fig:att2}}
    \caption{Attention distribution of the 5th block for SUBLLM and the 6th block for LLaMA, where kept indexes in subsampling are highlighted in red.}
    \vspace{5mm}
    \label{fig:attention}
\end{figure}

To analyze the distribution of indexes after subsampling, we examine the attention distribution of the 1.3B SUBLLM model retaining 40\% of tokens with two subsampling modules. 
First, we compare the index distribution after the first subsampling in SUBLLM model with the attention distribution within the pre-subsampling block (the fifth block). 
As illustrated in Figure~\ref{fig:att1}, it is evident that most tokens which receive significant attention, visible as distinct vertical stripes in the pre-subsampling attention distribution, are preserved by the subsampling module. 
Additionally, we analyze attention distribution of the sixth block in the 1.3B LLaMA model, where SUBLLM begins to compute on its shorter sequences after first subsampling.
We compare it with the same index retention distribution following the first subsampling module in SUBLLM. 
In this study, we hypothesize that for language models, the semantics at equivalent depths should be similar. 
As shown in Figure~\ref{fig:att2}, it can be observed that the tokens identified as crucial by LLaMA align closely with the subsampled regions where attention is computed in SUBLLM at the same depth, which further indicates effective preservation of important semantic information through the subsampling process.

\section{Conclusion}
In this study, we propose SUBLLM, a novel network architecture that utilizes text sequence redundancy and token significance to enhance training and decoding speeds while preserving few-shot learning capabilities. SUBLLM features an innovative subsampling mechanism allowing for customizable token retention ratios and includes a bypass module that significantly speeds up model convergence. Our findings indicate that the ScaledAdam optimizer supports this architecture by enhancing its convergence performance. This architecture is compatible with existing optimization methods within the LLaMA model family, ensuring wide applicability. Future research will investigate the impact of sequence compression ratios on SUBLLM to further understand token sequence subsampling, as well as further validate the model's scalability.

\bibliography{mybibfile}

\begin{thebibliography}{49}
\providecommand{\natexlab}[1]{#1}
\providecommand{\url}[1]{\texttt{#1}}
\expandafter\ifx\csname urlstyle\endcsname\relax
  \providecommand{\doi}[1]{doi: #1}\else
  \providecommand{\doi}{doi: \begingroup \urlstyle{rm}\Url}\fi

\bibitem[Ainslie et~al.(2023)Ainslie, Lei, de~Jong, Onta{\~{n}}{\'{o}}n, Brahma, Zemlyanskiy, et~al.]{ainslie2023colt5}
J.~Ainslie, T.~Lei, M.~de~Jong, S.~Onta{\~{n}}{\'{o}}n, S.~Brahma, Y.~Zemlyanskiy, et~al.
\newblock Colt5: Faster long-range transformers with conditional computation.
\newblock In \emph{Proceedings of EMNLP}, pages 5085--5100, 2023.

\bibitem[Bapna et~al.(2020)Bapna, Arivazhagan, and Firat]{bapna2020controlling}
A.~Bapna, N.~Arivazhagan, and O.~Firat.
\newblock Controlling computation versus quality for neural sequence models.
\newblock \emph{arXiv preprint arXiv:2002.07106}, 2020.

\bibitem[Barbieri et~al.(2020)Barbieri, Camacho{-}Collados, Anke, and Neves]{barbieri2020tweeteval}
F.~Barbieri, J.~Camacho{-}Collados, L.~E. Anke, and L.~Neves.
\newblock Tweeteval: Unified benchmark and comparative evaluation for tweet classification.
\newblock In \emph{Findings of EMNLP 2020}, pages 1644--1650, 2020.

\bibitem[Botev et~al.(2024)Botev, De, Smith, Fernando, Muraru, et~al.]{botev2024recurrentgemma}
A.~Botev, S.~De, S.~L. Smith, A.~Fernando, G.-C. Muraru, et~al.
\newblock Recurrentgemma: Moving past transformers for efficient open language models.
\newblock \emph{arXiv preprint arXiv:2404.07839}, 2024.

\bibitem[Brown et~al.(2020)Brown, Mann, Subbiah, et~al.]{brown2020language}
T.~B. Brown, B.~Mann, M.~Subbiah, et~al.
\newblock Language models are few-shot learners.
\newblock \emph{Advances in Neural Information Processing Systems}, 33:\penalty0 1877--1901, 2020.

\bibitem[Cai et~al.(2024)Cai, Li, Geng, Peng, Lee, Chen, and Dao]{cai2024medusa}
T.~Cai, Y.~Li, Z.~Geng, H.~Peng, J.~D. Lee, D.~Chen, and T.~Dao.
\newblock Medusa: Simple llm inference acceleration framework with multiple decoding heads.
\newblock \emph{arXiv preprint arXiv:2401.10774}, 2024.

\bibitem[Dai et~al.(2020)Dai, Lai, Yang, and Le]{dai2020funnel}
Z.~Dai, G.~Lai, Y.~Yang, and Q.~Le.
\newblock Funnel-transformer: Filtering out sequential redundancy for efficient language processing.
\newblock \emph{Advances in Neural Information Processing Systems}, 33:\penalty0 4271--4282, 2020.

\bibitem[Dao(2023)]{dao2023flashattention}
T.~Dao.
\newblock Flashattention-2: Faster attention with better parallelism and work partitioning.
\newblock \emph{arXiv preprint arXiv:2307.08691}, 2023.

\bibitem[Frantar and Alistarh(2023)]{frantar2023sparsegpt}
E.~Frantar and D.~Alistarh.
\newblock Sparsegpt: Massive language models can be accurately pruned in one-shot.
\newblock In \emph{International Conference on Machine Learning}, pages 10323--10337. PMLR, 2023.

\bibitem[Frantar et~al.(2022)Frantar, Ashkboos, Hoefler, and Alistarh]{frantar2022gptq}
E.~Frantar, S.~Ashkboos, T.~Hoefler, and D.~Alistarh.
\newblock Gptq: Accurate post-training quantization for generative pre-trained transformers.
\newblock \emph{arXiv preprint arXiv:2210.17323}, 2022.

\bibitem[Gu and Dao(2023)]{gu2023mamba}
A.~Gu and T.~Dao.
\newblock Mamba: Linear-time sequence modeling with selective state spaces.
\newblock \emph{arXiv preprint arXiv:2312.00752}, 2023.

\bibitem[He et~al.(2023)He, Yang, Feng, Yin, Wang, Leng, and Lin]{he2023fourier}
Z.~He, M.~Yang, M.~Feng, J.~Yin, X.~Wang, J.~Leng, and Z.~Lin.
\newblock Fourier transformer: Fast long range modeling by removing sequence redundancy with fft operator.
\newblock In \emph{Findings of ACL 2023}, pages 8954--8966, 2023.

\bibitem[Hendrycks et~al.(2021)Hendrycks, Burns, Basart, Zou, Mazeika, Song, and Steinhardt]{mmlu2021}
D.~Hendrycks, C.~Burns, S.~Basart, A.~Zou, M.~Mazeika, D.~Song, and J.~Steinhardt.
\newblock Measuring massive multitask language understanding.
\newblock \emph{Proceedings of ICLR}, 2021.

\bibitem[Hinton et~al.(2015)Hinton, Vinyals, and Dean]{hinton2015distilling}
G.~E. Hinton, O.~Vinyals, and J.~Dean.
\newblock Distilling the knowledge in a neural network.
\newblock \emph{arXiv preprint arXiv:1503.02531}, 2015.

\bibitem[Jiang et~al.(2024)Jiang, Li, Zhang, Wu, Luo, Ahn, Han, et~al.]{jiang2024minference}
H.~Jiang, Y.~Li, C.~Zhang, Q.~Wu, X.~Luo, S.~Ahn, Z.~Han, et~al.
\newblock Minference 1.0: Accelerating pre-filling for long-context llms via dynamic sparse attention.
\newblock \emph{arXiv preprint arXiv:2407.02490}, 2024.

\bibitem[Lehmann et~al.(2015)Lehmann, Isele, Jakob, Jentzsch, Kontokostas, Mendes, Hellmann, Morsey, van Kleef, Auer, and Bizer]{lehmann2015dbpedia}
J.~Lehmann, R.~Isele, M.~Jakob, A.~Jentzsch, D.~Kontokostas, P.~N. Mendes, S.~Hellmann, M.~Morsey, P.~van Kleef, S.~Auer, and C.~Bizer.
\newblock Dbpedia--a large-scale, multilingual knowledge base extracted from wikipedia.
\newblock \emph{Semantic Web}, 6\penalty0 (2):\penalty0 167--195, 2015.

\bibitem[Lehmann et~al.(2024)Lehmann, Isele, Jakob, Jentzsch, Kontokostas, Mendes, Hellmann, Morsey, van Kleef, et~al.]{liu2023scissorhands}
J.~Lehmann, R.~Isele, M.~Jakob, A.~Jentzsch, D.~Kontokostas, P.~N. Mendes, S.~Hellmann, M.~Morsey, P.~van Kleef, et~al.
\newblock Dbpedia - a large-scale, multilingual knowledge base extracted from wikipedia.
\newblock \emph{Advances in Neural Information Processing Systems}, 36, 2024.

\bibitem[Lei et~al.(2023)Lei, Bai, Brahma, Ainslie, Lee, Zhou, Du, et~al.]{NEURIPS2023_19d7204a}
T.~Lei, J.~Bai, S.~Brahma, J.~Ainslie, K.~Lee, Y.~Zhou, N.~Du, et~al.
\newblock Conditional adapters: Parameter-efficient transfer learning with fast inference.
\newblock \emph{Advances in Neural Information Processing Systems}, 36:\penalty0 8152--8172, 2023.

\bibitem[Leviathan et~al.(2023)Leviathan, Kalman, and Matias]{leviathan2023fast}
Y.~Leviathan, M.~Kalman, and Y.~Matias.
\newblock Fast inference from transformers via speculative decoding.
\newblock In \emph{International Conference on Machine Learning}, pages 19274--19286. PMLR, 2023.

\bibitem[Ma et~al.(2024)Ma, Yang, Xiong, Chen, Yu, Zhang, May, Zettlemoyer, Levy, and Zhou]{ma2024megalodon}
X.~Ma, X.~Yang, W.~Xiong, B.~Chen, L.~Yu, H.~Zhang, J.~May, L.~S. Zettlemoyer, O.~Levy, and C.~Zhou.
\newblock Megalodon: Efficient llm pretraining and inference with unlimited context length.
\newblock \emph{arXiv preprint arXiv:2404.08801}, 2024.

\bibitem[Magister et~al.(2023)Magister, Mallinson, Ad{\'{a}}mek, Malmi, and Severyn]{magister2023teaching}
L.~C. Magister, J.~Mallinson, J.~Ad{\'{a}}mek, E.~Malmi, and A.~Severyn.
\newblock Teaching small language models to reason.
\newblock In \emph{Proceedings of ACL}, pages 1773--1781, 2023.

\bibitem[Miao et~al.(2023)Miao, Oliaro, Zhang, Cheng, Wang, Wong, Chen, Arfeen, Abhyankar, and Jia]{miao2023specinfer}
X.~Miao, G.~Oliaro, Z.~Zhang, X.~Cheng, Z.~Wang, R.~Y.~Y. Wong, Z.~Chen, D.~Arfeen, R.~Abhyankar, and Z.~Jia.
\newblock Specinfer: Accelerating generative llm serving with speculative inference and token tree verification.
\newblock \emph{arXiv preprint arXiv:2305.09781}, 2023.

\bibitem[Nawrot et~al.(2024)Nawrot, Łańcucki, Chochowski, Tarjan, and Ponti]{nawrot2024dmc}
P.~Nawrot, A.~Łańcucki, M.~Chochowski, D.~Tarjan, and E.~M. Ponti.
\newblock Dynamic memory compression: Retrofitting llms for accelerated inference.
\newblock \emph{arXiv preprint arXiv:2403.09636}, 2024.

\bibitem[Ott et~al.(2019)Ott, Edunov, Baevski, Fan, Gross, Ng, Grangier, and Auli]{ott2019fairseq}
M.~Ott, S.~Edunov, A.~Baevski, A.~Fan, S.~Gross, N.~Ng, D.~Grangier, and M.~Auli.
\newblock Fairseq: A fast, extensible toolkit for sequence modeling.
\newblock In \emph{Proceedings of NAACL-HLT 2019: Demonstrations}, 2019.

\bibitem[Peng et~al.(2023)Peng, Alcaide, Anthony, et~al.]{peng2023rwkv}
B.~Peng, E.~Alcaide, Q.~Anthony, et~al.
\newblock Rwkv: Reinventing rnns for the transformer era.
\newblock \emph{arXiv preprint arXiv:2305.13048}, 2023.

\bibitem[Raffel et~al.(2020)Raffel, Shazeer, Roberts, Lee, Narang, et~al.]{JMLR:v21:20-074}
C.~Raffel, N.~Shazeer, A.~Roberts, K.~Lee, S.~Narang, et~al.
\newblock Exploring the limits of transfer learning with a unified text-to-text transformer.
\newblock \emph{Journal of Machine Learning Research}, 21\penalty0 (140):\penalty0 1--67, 2020.

\bibitem[Rajbhandari et~al.(2020)Rajbhandari, Rasley, Ruwase, and He]{rajbhandari2020zero}
S.~Rajbhandari, J.~Rasley, O.~Ruwase, and Y.~He.
\newblock Zero: Memory optimizations toward training trillion parameter models.
\newblock In \emph{SC20: International Conference for High Performance Computing, Networking, Storage and Analysis}, pages 1--16. IEEE, 2020.

\bibitem[Raposo et~al.(2024)Raposo, Ritter, Richards, Lillicrap, Humphreys, and Santoro]{raposo2024mixture}
D.~Raposo, S.~Ritter, B.~A. Richards, T.~P. Lillicrap, P.~C. Humphreys, and A.~Santoro.
\newblock Mixture-of-depths: Dynamically allocating compute in transformer-based language models.
\newblock \emph{arXiv preprint arXiv:2404.02258}, 2024.

\bibitem[Shen et~al.(2024)Shen, Lin, Zha, Liu, Luan, Wang, and Wang]{shen-etal-2024-pruning}
B.~Shen, Z.~Lin, D.~Zha, W.~Liu, J.~Luan, B.~Wang, and W.~Wang.
\newblock Pruning large language models to intra-module low-rank architecture with transitional activations.
\newblock In \emph{Findings of ACL 2024}, pages 9781--9793, 2024.

\bibitem[Soboleva et~al.(2023)Soboleva, Al-Khateeb, Myers, Steeves, Hestness, and Dey]{cerebras2023slimpajama}
D.~Soboleva, F.~Al-Khateeb, R.~Myers, J.~R. Steeves, J.~Hestness, and N.~Dey.
\newblock Slimpajama: A 627b token cleaned and deduplicated version of redpajama.
\newblock \emph{https://www.cerebras.net/blog/slimpajama-a-627b-token-cleaned-and-deduplicated-version-of-redpajama}, June 2023.

\bibitem[Socher et~al.(2013)Socher, Perelygin, Wu, Chuang, Manning, Ng, and Potts]{socher2013recursive}
R.~Socher, A.~Perelygin, J.~Wu, J.~Chuang, C.~D. Manning, A.~Y. Ng, and C.~Potts.
\newblock Recursive deep models for semantic compositionality over a sentiment treebank.
\newblock In \emph{Proceedings of EMNLP}, pages 1631--1642, 2013.

\bibitem[Spector and R{\'{e}}(2023)]{spector2023accelerating}
B.~Spector and C.~R{\'{e}}.
\newblock Accelerating llm inference with staged speculative decoding.
\newblock \emph{arXiv preprint arXiv:2308.04623}, 2023.

\bibitem[Su et~al.(2024)Su, Ahmed, Lu, Pan, Bo, and Liu]{su2024roformer}
J.~Su, M.~H.~M. Ahmed, Y.~Lu, S.~Pan, W.~Bo, and Y.~Liu.
\newblock Roformer: Enhanced transformer with rotary position embedding.
\newblock \emph{Neurocomputing}, 568:\penalty0 127063, 2024.

\bibitem[Sun et~al.(2024)Sun, Dong, Zhu, Huang, Wang, Ma, Zhang, Wang, and Wei]{sun2024yoco}
Y.~Sun, L.~Dong, Y.~Zhu, S.~Huang, W.~Wang, S.~Ma, Q.~Zhang, J.~Wang, and F.~Wei.
\newblock You only cache once: Decoder-decoder architectures for language models.
\newblock \emph{arXiv preprint arXiv:2405.05254}, 2024.

\bibitem[Suzgun et~al.(2022)Suzgun, Scales, Sch{\"{a}}rli, Gehrmann, Tay, Chung, Chowdhery, Le, Chi, Zhou, and Wei]{suzgun2022bbh}
M.~Suzgun, N.~Scales, N.~Sch{\"{a}}rli, S.~Gehrmann, Y.~Tay, H.~W. Chung, A.~Chowdhery, Q.~V. Le, E.~H. Chi, D.~Zhou, and J.~Wei.
\newblock Challenging big-bench tasks and whether chain-of-thought can solve them.
\newblock \emph{arXiv preprint arXiv:2210.09261}, 2022.

\bibitem[Tay et~al.(2022)Tay, Dehghani, Bahri, and Metzler]{tay2022efficient}
Y.~Tay, M.~Dehghani, D.~Bahri, and D.~Metzler.
\newblock Efficient transformers: A survey.
\newblock \emph{ACM Computing Surveys}, 55\penalty0 (6):\penalty0 1--28, 2022.

\bibitem[Touvron et~al.(2023{\natexlab{a}})Touvron, Lavril, Izacard, Martinet, et~al.]{touvron2023llama}
H.~Touvron, T.~Lavril, G.~Izacard, X.~Martinet, et~al.
\newblock Llama: Open and efficient foundation language models.
\newblock \emph{arXiv preprint arXiv:2302.13971}, 2023{\natexlab{a}}.

\bibitem[Touvron et~al.(2023{\natexlab{b}})Touvron, Martin, Stone, Albert, Almahairi, Babaei, Bashlykov, et~al.]{touvron2023llama2}
H.~Touvron, L.~Martin, K.~Stone, P.~Albert, A.~Almahairi, Y.~Babaei, N.~Bashlykov, et~al.
\newblock Llama 2: Open foundation and fine-tuned chat models.
\newblock \emph{arXiv preprint arXiv:2307.09288}, 2023{\natexlab{b}}.

\bibitem[Wang et~al.(2024)Wang, Yuan, Yang, Zhang, Zhao, Liu, Luan, Povey, and Wang]{wang2024subllm}
Q.~Wang, Y.~Yuan, X.~Yang, R.~Zhang, K.~Zhao, W.~Liu, J.~Luan, D.~Povey, and B.~Wang.
\newblock Subllm: A novel efficient architecture with token sequence subsampling for llm.
\newblock \emph{arXiv preprint arXiv:2406.06571}, 2024.
\newblock Full version of this paper.

\bibitem[Wang et~al.(2020)Wang, Li, Khabsa, Fang, and Ma]{wang2020linformer}
S.~Wang, B.~Z. Li, M.~Khabsa, H.~Fang, and H.~Ma.
\newblock Linformer: Self-attention with linear complexity.
\newblock \emph{arXiv preprint arXiv:2006.04768}, 2020.

\bibitem[Xia et~al.(2024)Xia, Gao, Zeng, and Chen]{xia2024shearedllama}
M.~Xia, T.~Gao, Z.~Zeng, and D.~Chen.
\newblock Sheared llama: Accelerating language model pre-training via structured pruning.
\newblock \emph{arXiv preprint arXiv:2310.06694}, 2024.

\bibitem[Xiao et~al.(2023)Xiao, Lin, Seznec, et~al.]{xiao2023smoothquant}
G.~Xiao, J.~Lin, M.~Seznec, et~al.
\newblock Smoothquant: Accurate and efficient post-training quantization for large language models.
\newblock In \emph{International Conference on Machine Learning}, pages 38087--38099. PMLR, 2023.

\bibitem[Yang et~al.(2023)Yang, Ge, Wang, Jiao, Jiang, Yang, Majumder, and Wei]{yang2023inference}
N.~Yang, T.~Ge, L.~Wang, B.~Jiao, D.~Jiang, L.~Yang, R.~Majumder, and F.~Wei.
\newblock Inference with reference: Lossless acceleration of large language models.
\newblock \emph{arXiv preprint arXiv:2304.04487}, 2023.

\bibitem[Yao et~al.(2023)Yao, Guo, Yang, Kang, Kuang, Yang, Jin, Lin, and Povey]{yao2023zipformer}
Z.~Yao, L.~Guo, X.~Yang, W.~Kang, F.~Kuang, Y.~Yang, Z.~Jin, L.~Lin, and D.~Povey.
\newblock Zipformer: A faster and better encoder for automatic speech recognition.
\newblock \emph{arXiv preprint arXiv:2310.11230}, 2023.

\bibitem[Zhai et~al.(2021)Zhai, Talbott, Srivastava, Huang, Goh, Zhang, and Susskind]{zhai2021attention}
S.~Zhai, W.~Talbott, N.~Srivastava, C.~Huang, H.~Goh, R.~Zhang, and J.~M. Susskind.
\newblock An attention free transformer.
\newblock \emph{arXiv preprint arXiv:2105.14103}, 2021.

\bibitem[Zhang et~al.(2015)Zhang, Zhao, and LeCun]{zhang2015character}
X.~Zhang, J.~J. Zhao, and Y.~LeCun.
\newblock Character-level convolutional networks for text classification.
\newblock \emph{Advances in Neural Information Processing Systems}, 28, 2015.

\bibitem[Zhang et~al.(2023)Zhang, Sheng, Zhou, Chen, Zheng, Cai, Song, Tian, Re, Barrett, Wang, and Chen]{zhang2023ho}
Z.~Zhang, Y.~Sheng, T.~Zhou, T.~Chen, L.~Zheng, R.~Cai, Z.~Song, Y.~Tian, C.~Re, C.~Barrett, Z.~Wang, and B.~Chen.
\newblock H2o: Heavy-hitter oracle for efficient generative inference of large language models.
\newblock \emph{Advances in Neural Information Processing Systems}, 36, 2023.

\bibitem[Zhong et~al.(2023)Zhong, Cui, Guo, Liang, Lu, Wang, Saied, Chen, and Duan]{zhong2023agieval}
W.~Zhong, R.~Cui, Y.~Guo, Y.~Liang, S.~Lu, Y.~Wang, A.~Saied, W.~Chen, and N.~Duan.
\newblock Agieval: A human-centric benchmark for evaluating foundation models.
\newblock \emph{arXiv preprint arXiv:2304.06364}, 2023.

\bibitem[Zhu et~al.(2023)Zhu, Li, Liu, Ma, and Wang]{zhu2023survey}
X.~Zhu, J.~Li, Y.~Liu, C.~Ma, and W.~Wang.
\newblock A survey on model compression for large language models.
\newblock \emph{arXiv preprint arXiv:2308.07633}, 2023.

\end{thebibliography}

\appendix 
\appendix 
\section{Training Details}
\label{sec:training_details}

Table \ref{tab:SUBLLM_structure} illustrates the structures of the SUBLLM models used in the experiments. The trained 0.25B models corresponds to the second row, which has a 15-block configuration, while the 1.3B model corresponds to the third row, which has a 24-block configuration. Both models utilize two subsampling modules with each retention ratio of approximately 63.24\%, dropping to a minimum of 40\% midway through the blocks. Table \ref{tab:model_config} lists the configurations for the LLaMA and SUBLLM model architectures.

The 1.3B model is trained on four nodes of 8 A100 GPUs, each with 80GB memory, while the 0.25B model is trained on one node.  We use ScaledAdam optimizer with $\beta_1 = 0.9$, $\beta_2 = 0.95$, $\epsilon=10^{-8}$.
At the beginning of training, we set $c_{min}$ and $c_{max}$ in the bypass module to 0.9 and 1.0 respectively. Then, we gradually decrease $c_{min}$ to 0.2 after 20k steps. The inference is performed on one GPU of A100. 

\begin{table}[htbp]
\centering
\caption{Model configurations of 1.3B and 0.25B models.}
\vspace{4mm}
\label{tab:model_config}
\begin{tabular}{lcc}
\toprule
\textbf{Configuration} & \textbf{0.25B} & \textbf{1.3B} \\ \midrule
num\_hidden\_layers    & 15             & 24            \\
hidden\_size           & 1024           & 2048          \\
intermediate\_size     & 4096           & 5504          \\
num\_attention\_heads  & 16             & 16            \\
tie\_word\_embeddings  & false          & false         \\
rope\_theta            & 10000.0        & 10000.0       \\ \bottomrule
\end{tabular}
\end{table}

\section{Module Details}
\begin{figure}[htbp]
    \centering
    \includegraphics[width=0.4\textwidth]{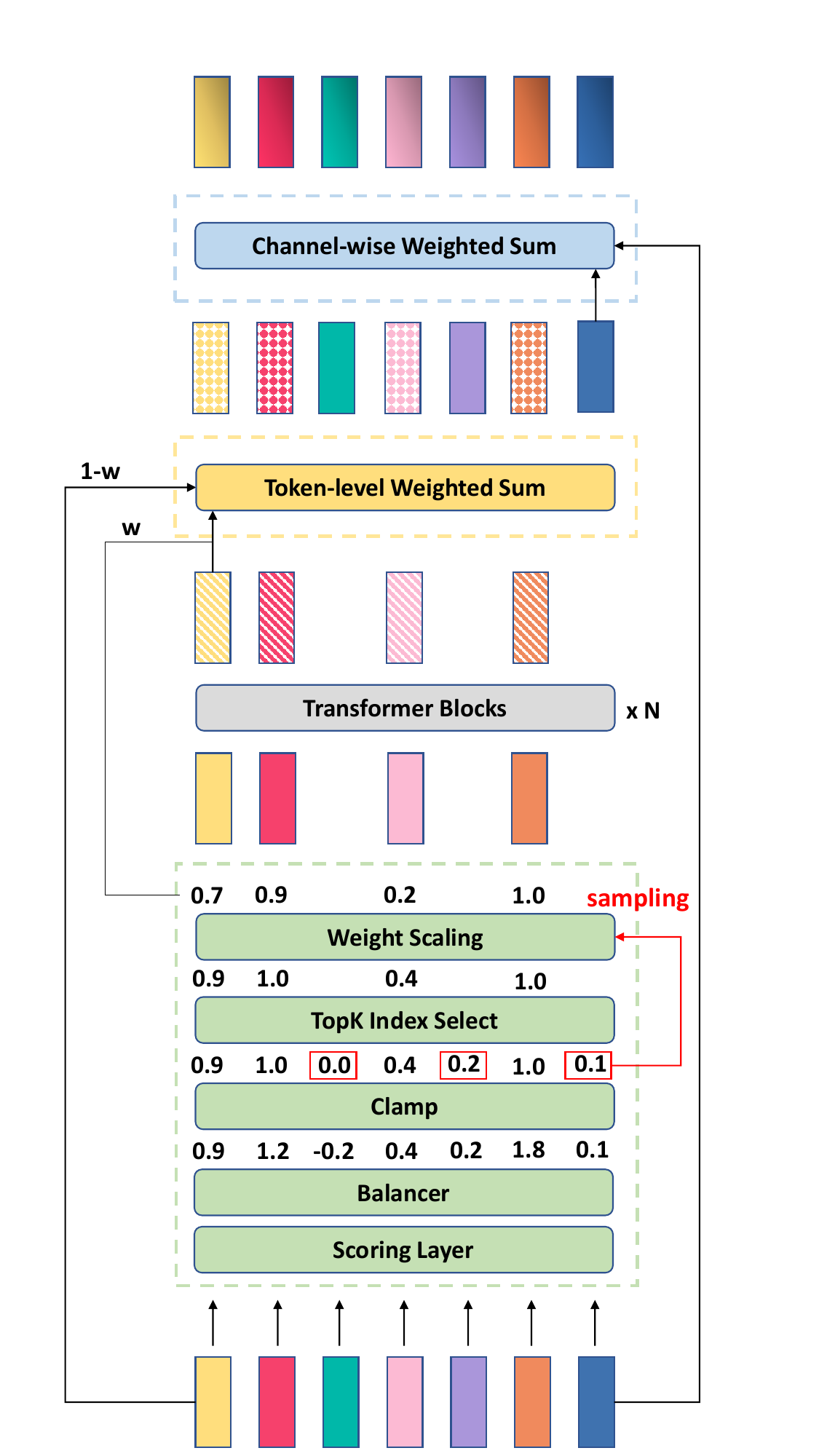}
    \caption{A demonstration of the computation of subsampling, upsampling and bypass modules. From bottom to top, the first dashed box represents the learnable subsampling module, the second dashed box represents the upsampling module, and the third dashed box represents the bypass module. \textcolor{red}{Sampling} denotes performing sampling with replacement from the weights of the tokens not selected, and these weights are used to be subtracted from the weights of the selected tokens.}
    \label{fig:module_details}
\end{figure}

Figure \ref{fig:module_details} illustrates the computation process of subsampling, upsampling, and bypass. To enhance readability and facilitate understanding, we have consolidated the operations from Equation \ref{eq:index_select} to Equation \ref{eq:weight_scaling} within the subsampling module.

\section{Model Performance}

\begin{figure}[htbp]
    \centering
    \includegraphics[width=0.4\textwidth]{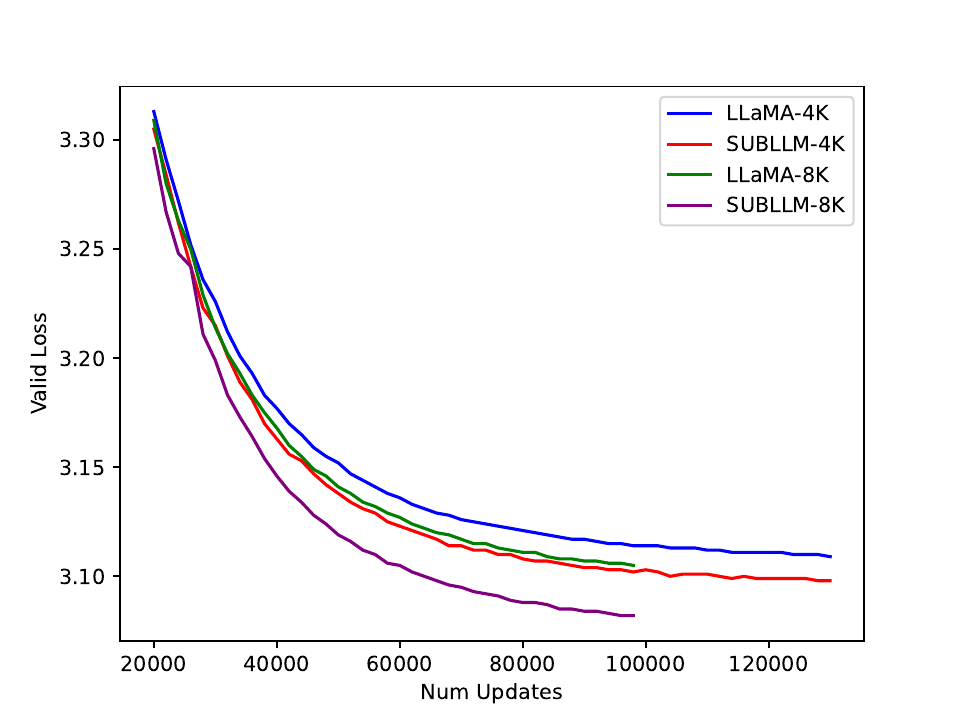}
    \caption{Valid loss comparison of 1.3B models under 4K and 8K Settings.}
    \label{fig:valid_loss_1p3b}
\end{figure}

\paragraph{Performance on Benchmarks} We report results for 1.3B SUBLLM and LLaMA models with 4k context length across three benchmarks, MMLU (5-shot), BBH (3-shot) and AGIEval (5-shot). As shown in Table \ref{tab:benchmark_results}, the results of both models on the benchmarks are close to random guessing due to the insufficient amount of training data. However, SUBLLM performs slightly better than LLaMA.

\begin{table}[htb]
    \centering
    \caption{Results for 1.3B SUBLLM and LLaMA models with 4K context length across three benchmarks.}
    \vspace{5mm}
    \begin{tabular}{lccc}
    \toprule
    Model  & MMLU           & BBH            & AGIEval        \\ \midrule
    LLaMA  & 26.23          & 23.70          & 16.76          \\
    SUBLLM & \textbf{26.41} & \textbf{24.17} & \textbf{17.64} \\ \bottomrule
    \end{tabular}
    \label{tab:benchmark_results}
\end{table}

\paragraph{Results on Valid Loss} We report the valid loss of the 1.3B SUBLLM and LLaMA models trained with 4k and 8k context lengths. 
The SUBLLM is configured with twice subsampling and a total retention ratio of 40\%. 
As shown in Figure \ref{fig:valid_loss_1p3b}, the valid loss of SUBLLM is significantly lower than that of LLaMA under different context window lengths. This indicated the potential of our proposed model architecture SUBLLM in accelerating training while maintaining model performance. The gap in valid loss between LLaMA and SUBLLM widens as the training context window length increases. This might imply the potential of SUBLLM in handling long sequences. We leave this exploration for future work.

\section{Limitation}
Our work faced two primary limitations that affected the scalability of our proposed architecture. First, due to limited computational resources, we were only able to train models with a maximum parameter size of 1.3 billion. This restriction prevented us from exploring the full potential of larger models, which could offer improved performance on more complex tasks. Second, the amount of data we used for training was also constrained. In our experiments with the 0.25B model, we found that using data equivalent to one hundred times the model's parameters was sufficient for convergence. We applied the same approach to the 1.3B models, which meant we could not leverage the entire Slimpajama dataset. Together, these limitations hindered our ability to fully assess the model's scalability. In future work, we will focus on validating the scalability of the proposed model architecture with larger models and more extensive datasets.

\end{document}